\documentclass[journal]{IEEEtran}
\usepackage{graphicx}
\usepackage{amsmath}
\usepackage{amssymb}
\usepackage{booktabs}
\usepackage{hyperref}
\usepackage{cite}
\usepackage{xcolor}
\usepackage{url}

\begin{document}

\title{GlacierCastAI: Predicting Glacier Retreat
from Multi-Modal Satellite Imagery and Climate Signals}

\author{Arunkumar Ramachandran\\
Independent Researcher\\
\texttt{arunkumar.ramachandran.research@gmail.com}}

\maketitle

\begin{abstract}
ERA5 seasonal climate variables contain predictive information
about future glacier retreat that extends beyond what satellite
imagery alone provides --- yet existing deep learning methods
focus on mapping current boundaries rather than forecasting
future ones. This paper presents \textbf{GlacierCastAI},
which reframes glacier boundary prediction as a multi-modal
spatiotemporal forecasting problem rather than a mapping
problem, fusing multi-temporal Landsat imagery with ERA5
reanalysis climate variables and Copernicus DEM terrain
features to forecast glacier boundaries across five study
glaciers spanning four climate regimes. The architecture
couples a ResNet50 spatial encoder with a ConvLSTM temporal
model and a cross-attention climate fusion module. Because
forecasting future boundaries from historical sequences is
inherently more uncertain than delineating current boundaries
from high-quality imagery, the reported IoU values
(0.320--0.337) are not directly comparable to those of
state-of-the-art mapping models. The relevant comparisons
are against traditional baselines and between experimental
conditions. Through a pre-registered ablation study, adding
ERA5 climate signals is shown to improve image-only IoU from
0.326 to 0.337 (+3.4\%), suggesting that atmospheric forcing
carries predictive information beyond what imagery alone
provides. All deep learning models substantially outperform
traditional persistence and linear trend baselines
(IoU 0.160 and 0.169 respectively), with improvements of
89--99\% relative IoU. A lightweight climate-only MLP
baseline (661K parameters) achieves an IoU of 0.320---98\%
of image-only performance---using 85$\times$ fewer
parameters, suggesting that ERA5 variables encode substantial
predictive signal independently of satellite imagery. SHAP
attribution analysis suggests that spring solar radiation
(MAM) is the dominant climate driver, consistent with the
known role of spring insolation in setting melt season
trajectories. Code and results are available at
\url{https://github.com/Arun-K-Ram/GlacierCastAI}.
\end{abstract}


\section{Introduction}
\label{sec:intro}

Glaciers cover approximately 170,000~km$^2$ of Earth's land
surface outside the ice sheets and represent critical freshwater
reserves for over two billion people~\cite{rgi2017}. Global
glacier mass loss has accelerated markedly since 2000, with
Hugonnet et al.\ reporting an average loss of
$267 \pm 16$~Gt~yr$^{-1}$ between 2000 and 2019~\cite{hugonnet2021}.
This retreat threatens water security, raises sea levels, and
destabilizes mountain ecosystems. Early warning systems capable
of predicting retreat trajectories years in advance are therefore
essential for infrastructure planning and climate adaptation.

Existing remote sensing approaches primarily address glacier
\textit{mapping}---delineating current boundaries from optical
imagery using spectral indices such as the Normalized Difference
Snow Index (NDSI)~\cite{ndsi1994} or deep segmentation
models~\cite{glavitu2024}. Although these methods have reached a high level of maturity, their focus remains on characterizing the current glacier extent, rather than addressing the operationally critical challenge of predicting glacier evolution over the next several years.

A key motivation for climate-augmented forecasting is the
temporal lag between climate forcing and visible glacier
response. Rising summer temperatures drive subsurface melting
and dynamic instability months to years before the glacier
terminus visibly retreats. Bolibar et al.\ demonstrated that
deep learning captures nonlinear sensitivities of glacier mass
balance to temperature and precipitation that linear models
miss~\cite{bolibar2022}, yet their approach targets scalar
mass balance estimates rather than spatially explicit boundary
forecasts. Existing work has focused primarily on current
boundary delineation or scalar mass balance projection,
whereas this work reframes the problem as multi-modal
spatiotemporal forecasting and empirically tests whether
climate signals provide predictive lead over imagery at
the patch level.

The primary research question addressed in this study is whether climate signals can predict glacier retreat before it becomes detectable in satellite imagery.

To address this, \textbf{GlacierCastAI} is introduced, a
spatiotemporal forecasting system that fuses three modalities:
(i) multi-temporal Landsat surface reflectance sequences
spanning 2000--2023; (ii) ERA5 reanalysis seasonal climate
variables (temperature, precipitation, snowfall, solar
radiation)~\cite{era5}; and (iii) Copernicus DEM GLO-30
terrain features (elevation, slope, aspect).

The contributions of this work are:
\begin{enumerate}
    \item \textbf{Problem reframing}: casting glacier boundary
          prediction as a multi-modal spatiotemporal
          \textit{forecasting} problem rather than a
          \textit{mapping} problem, and constructing the
          first dataset and evaluation protocol for this
          task across five climatically diverse glaciers.
    \item A pre-registered ablation study isolating the
          marginal contribution of each modality (imagery,
          climate, terrain) to forecast accuracy, with
          traditional persistence and linear trend baselines
          for reference.
    \item Empirical evidence that a climate-only MLP
          (661K parameters) achieves 98\% of image-only
          performance, suggesting the predictive
          sufficiency of ERA5 signals for glacier boundary
          forecasting.
    \item SHAP attribution analysis identifying spring
          solar radiation as the dominant climate driver
          of predicted retreat, providing physically
          interpretable early warning signals.
    \item A reproducible codebase and per-experiment result
          registry enabling transparent comparison with
          future work.
\end{enumerate}

\begin{figure}[t]
\centering
\includegraphics[width=\columnwidth]{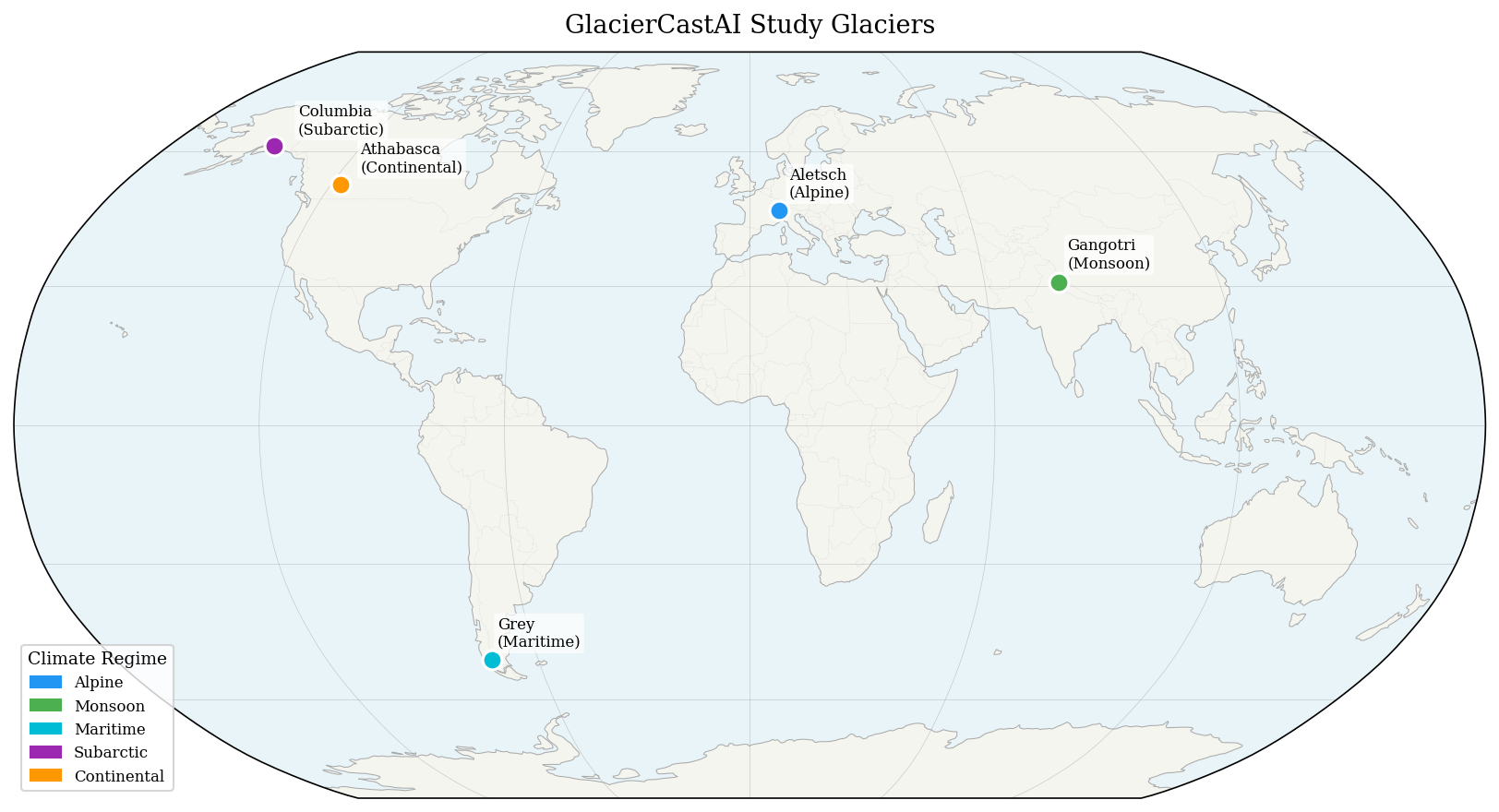}
\caption{Study glaciers across five climate regimes. Colors
denote climate regime: Alpine (blue), Monsoon (green),
Maritime (cyan), Maritime/Subarctic (purple), Continental
(orange).}
\label{fig:study_area}
\end{figure}

\section{Related Work}
\label{sec:related}

\subsection{Glacier Mapping from Satellite Imagery}
Automated glacier delineation has been studied extensively
using multispectral imagery. Band ratio methods such as NDSI
provide reliable ice/snow discrimination under clear-sky
conditions~\cite{ndsi1994}. Deep learning approaches including
U-Net~\cite{unet2015} variants have improved boundary
precision, particularly for debris-covered glaciers. Most
recently, Maslov et al.\ proposed GlaViTU, a hybrid
convolutional-transformer model achieving IoU above 0.85
on unseen imagery across multiple regions~\cite{glavitu2024}.
Importantly, these methods target \textit{current} boundary
delineation; GlacierCastAI targets \textit{future} boundary
forecasting, a fundamentally different task with inherently
higher uncertainty that is not directly comparable to
mapping benchmarks.

\subsection{Glacier Change and Mass Balance Modelling}
Multi-temporal Landsat analysis has quantified global and
regional retreat rates over four decades~\cite{zemp2019,hugonnet2021}.
Physics-based models such as the Open Global Glacier
Model (OGGM)~\cite{maussion2019} simulate glacier evolution
using climate forcing and ice flow dynamics, but require
detailed calibration data unavailable at scale. Bolibar
et al.\ applied deep learning to glacier mass balance
projection, revealing nonlinear sensitivities to temperature
and precipitation that linear temperature-index models
underestimate~\cite{bolibar2022}. The present work
complements theirs by targeting spatially explicit boundary
forecasting rather than scalar mass balance.

\subsection{Spatiotemporal Deep Learning for Earth Observation}
ConvLSTM~\cite{convlstm2015} introduced recurrent convolutional
architectures for spatiotemporal prediction, originally
applied to precipitation nowcasting. Spatiotemporal
forecasting with satellite sequences has since been applied
to sea ice prediction and vegetation change. There is limited
prior work applying multi-modal spatiotemporal deep learning
to glacier boundary forecasting, and no prior study has
empirically quantified the predictive contribution of
climate reanalysis signals at the patch level.

\subsection{Explainability in Earth Observation}
SHAP (SHapley Additive exPlanations)~\cite{shap2017} provides
model-agnostic attribution of feature contributions to
individual predictions. Applied to climate-driven Earth
observation models, SHAP can identify which atmospheric
variables and seasons most strongly influence predictions,
connecting data-driven results to physical understanding.
This work applies SHAP to attribute the contribution of
each ERA5 seasonal variable to the predicted glacier boundary.

\section{Methodology}
\label{sec:method}

\subsection{Study Glaciers}
Five glaciers were intentionally selected to span diverse
climate regimes rather than maximize dataset size, enabling
evaluation of model generalization across Alpine,
monsoon-driven, maritime, subarctic, and continental
settings (Table~\ref{tab:glaciers}, Fig.~\ref{fig:study_area}).

\begin{table}[h]
\centering
\caption{Study Glaciers and Climate Regimes}
\label{tab:glaciers}
\begin{tabular}{llll}
\toprule
Glacier & Region & Climate Regime & Period \\
\midrule
Aletsch   & Swiss Alps       & Alpine             & 2002--2023 \\
Gangotri  & Himalayas        & Monsoon            & 2004--2023 \\
Grey      & Patagonia        & Maritime           & 2004--2023 \\
Columbia  & Alaska           & Maritime/Subarctic & 2001--2023 \\
Athabasca & Canadian Rockies & Continental        & 2001--2023 \\
\bottomrule
\end{tabular}
\end{table}

\subsection{Data Sources}

\subsubsection{Satellite Imagery}
Landsat Collection 2 Level-2 surface reflectance products
(Landsat 5, 7, 8, and 9) are used, accessed via the USGS
Earth Explorer and Microsoft Planetary Computer STAC API.
Summer acquisitions (June--September for Northern Hemisphere
glaciers; December--March for Grey Glacier, Patagonia) with
cloud cover below 20\% are selected. Band combinations
include Green (Band 3), NIR (Band 5), and SWIR1 (Band 6),
from which NDSI is derived as an auxiliary channel
($\text{NDSI} = (\text{Green} - \text{SWIR1}) / (\text{Green}
+ \text{SWIR1})$). The dataset spans 64 scenes across
all five glaciers covering 2000--2023.

\subsubsection{Climate Data}
ERA5 monthly mean reanalysis data from the Copernicus
Climate Data Store~\cite{era5} provides four variables
aggregated to seasonal means: 2m air temperature (t2m),
total precipitation (tp), snowfall (sf), and surface net
solar radiation (ssr). Four seasons (DJF, MAM, JJA, SON)
yield a 16-dimensional climate feature vector per timestep,
embedded as a time-varying auxiliary input to the temporal
model. Solar radiation values reaching approximately
$18 \times 10^6$~J~m$^{-2}$ are normalized using
physical-range standardization to prevent FP16 overflow
during mixed-precision training.

\subsubsection{Terrain Data}
The Copernicus DEM GLO-30 (30m resolution) provides
elevation data from which slope and aspect (decomposed
into sine and cosine components to avoid circular
discontinuity) are derived. DEM tiles are downloaded
via AWS open data, reprojected to UTM, and merged
per glacier region.

\subsection{Preprocessing Pipeline}
All raster inputs are co-registered to a common UTM
grid at 30m resolution. Landsat digital numbers are
converted to surface reflectance using Collection 2
scale factors ($\rho = \text{DN} \times 0.0000275 - 0.2$).
Glacier masks are derived by thresholding NDSI $> 0.4$.

Patches of $256 \times 256$ pixels are extracted using a
sliding window with 64-pixel overlap, retaining only patches
with at least 3\% glacier coverage and 70\% valid
(non-cloud) pixels. This yields 29,810 patches across
64 scenes. Temporal sequences of $T=4$ consecutive
timesteps are constructed per spatial location, paired
with target masks at the subsequent timestep. The dataset
comprises 40,476 sequences, split temporally to prevent
data leakage: test covers 2022--2023, validation covers
2016--2017, and training uses all prior years
(train: 27,725; val: 5,861; test: 6,890 sequences).

\subsection{Model Architecture}

\subsubsection{Spatial Encoder}
A ResNet50 backbone~\cite{resnet2016} pre-trained on
ImageNet encodes each of the $T=4$ timesteps independently,
producing spatial feature maps. The backbone is frozen
during the first five epochs (warmup) to prevent
randomly-initialized decoder weights from corrupting
pretrained representations, then fine-tuned end-to-end.

\subsubsection{Temporal Model and Climate Fusion}
Encoded spatial features from all timesteps are passed
to a ConvLSTM~\cite{convlstm2015} with three layers
(hidden dimension 256, kernel size $3 \times 3$).
Climate features are provided as a 16-dimensional
time-varying input injected at each ConvLSTM step via
cross-attention, allowing the model to condition spatial
predictions on atmospheric state. The fused spatiotemporal
representation is passed to all output heads.

\subsubsection{Output Heads}
Three output heads operate on the fused representation:
(i) a UNet-style~\cite{unet2015} decoder producing
per-pixel glacier probability maps (boundary mask head);
(ii) an MLP regression head predicting annual area loss
at 1, 2, and 3-year horizons (retreat rate head); and
(iii) a 3-class MLP classifier for accelerated retreat
risk (low/medium/high). In the ablation experiments,
the terrain branch (DEM projection layer) is activated
or deactivated to isolate its contribution.

\subsubsection{Climate-Only MLP Baseline}
To directly test whether climate signals alone carry
predictive information, a lightweight climate-only MLP
baseline is implemented. This model flattens the
$T \times F = 4 \times 16 = 64$-dimensional climate
sequence and passes it through a three-layer MLP
(hidden dimension 256), producing the same three
outputs as the full model. The boundary mask is
generated at $64 \times 64$ resolution and bilinearly
upsampled to $256 \times 256$. This model has 661K
parameters---approximately 85$\times$ fewer than the
56.1M-parameter full model---providing a parameter-efficient
lower bound for climate-driven forecasting.

\subsection{Loss Function}
The combined training objective is:
\begin{equation}
\mathcal{L} = \mathcal{L}_{\text{seg}} +
              \lambda_1 \mathcal{L}_{\text{retreat}} +
              \lambda_2 \mathcal{L}_{\text{risk}}
\label{eq:loss}
\end{equation}
where the segmentation loss combines Dice, binary
cross-entropy, and boundary-aware components:
\begin{equation}
\mathcal{L}_{\text{seg}} = 0.5\,\mathcal{L}_{\text{Dice}}
+ 0.3\,\mathcal{L}_{\text{BCE}}
+ 0.2\,\mathcal{L}_{\text{boundary}}
\end{equation}
The boundary loss upweights pixels within three pixels of
the glacier edge by a factor of $\theta = 19$, penalizing
coarse edge predictions. The retreat and risk loss weights
are $\lambda_1 = 0.5$ and $\lambda_2 = 0.3$.

\subsection{Training Details}
All models are trained with AdamW
($lr = 10^{-4}$, weight decay $= 10^{-4}$, $\beta = (0.9, 0.999)$)
with cosine annealing and 5-epoch linear warmup.
Mixed-precision training (FP16) is used throughout.
Early stopping monitors validation IoU with patience
of 15 epochs. Training is performed on an NVIDIA RTX 2060
GPU (6GB VRAM) with batch size 4. All experiments use
identical hyperparameters to ensure fair modality comparison.
Seeds are fixed at 42 for reproducibility. Results reflect
single runs per experimental condition; multi-seed evaluation
is an acknowledged limitation and is left for future work.

\subsection{SHAP Attribution}
To identify which ERA5 climate variables drive predicted
glacier retreat, SHAP KernelExplainer~\cite{shap2017}
is applied to the exp002 model. Imagery and terrain inputs
are held fixed at a reference sample while only the
16-dimensional climate feature vector is perturbed across
50 test patches, using 50 background samples for the
explainer and 100 perturbation samples per prediction.
SHAP values are averaged over the $T=4$ timesteps to
obtain a single attribution score per climate feature.

\section{Experiments}
\label{sec:experiments}

\subsection{Evaluation Metrics}
Two primary metrics are reported:
\begin{itemize}
    \item \textbf{IoU}: Intersection-over-Union of predicted
          versus ground-truth glacier mask, evaluated at
          a threshold of 0.5. Measures overall segmentation
          accuracy.
    \item \textbf{BF1}: Boundary F1, computed as the
          harmonic mean of precision and recall evaluated
          exclusively on glacier edge pixels (dilation width
          of 3 pixels). Captures boundary delineation
          quality independent of interior accuracy.
\end{itemize}

\subsection{Baselines}
The proposed method is compared against two traditional
forecasting baselines computed on the same test set:
\begin{itemize}
    \item \textbf{B1 --- Persistence}: predicts the next
          glacier boundary as identical to the current
          observed boundary (NDSI threshold $> 0.4$ on
          the last input timestep). Represents the
          null hypothesis that glaciers do not change.
    \item \textbf{B2 --- Linear trend}: fits a pixel-wise
          linear regression over the $T=4$ input NDSI
          values and extrapolates one step forward.
          Represents the assumption of steady linear retreat.
\end{itemize}

\subsection{Ablation Design}
Following the pre-registered protocol, four conditions
are compared, varying only the input modalities while
holding all other hyperparameters fixed:
\begin{itemize}
    \item \textbf{exp001 --- Image only}: imagery branch
          active; climate and DEM branches disabled.
    \item \textbf{exp002 --- Image + Climate}: imagery
          and ERA5 climate active; DEM disabled.
    \item \textbf{exp003 --- Image + Climate + DEM}:
          all three modalities active.
    \item \textbf{exp005 --- Climate only (MLP)}: lightweight
          MLP on ERA5 features only; no imagery or DEM.
\end{itemize}

Each experiment uses an isolated checkpoint directory
to prevent result contamination, verified before training
via directory inspection. Climate features are confirmed
non-zero via batch-level diagnostic prior to each run.

\section{Results}
\label{sec:results}

\begin{figure}[t]
\centering
\includegraphics[width=\columnwidth]{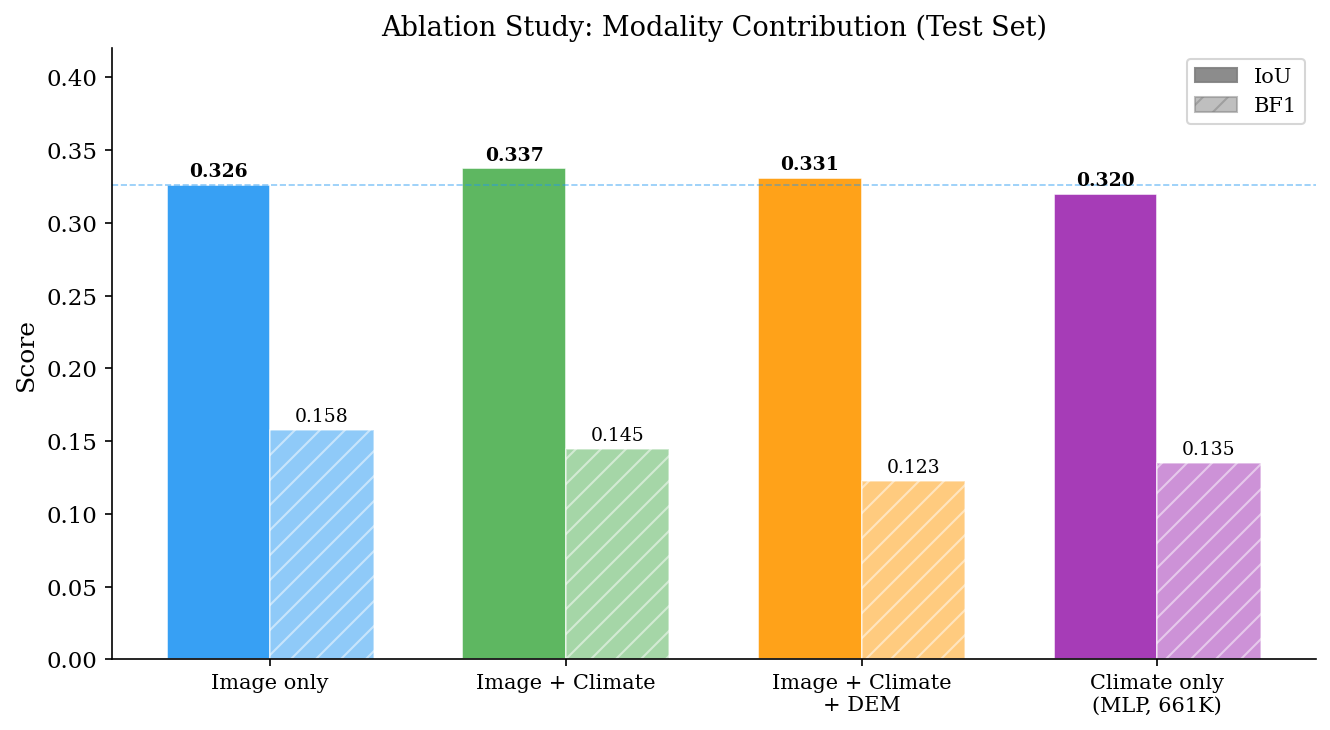}
\caption{IoU and BF1 across all experimental conditions.
Solid bars show IoU; hatched bars show BF1. Dashed line
indicates image-only IoU baseline (exp001).}
\label{fig:ablation}
\end{figure}

Table~\ref{tab:ablation} presents results on the held-out
test set (2022--2023), including traditional baselines.
Fig.~\ref{fig:ablation} visualizes the IoU and BF1
comparison across all conditions.

\begin{table}[h]
\centering
\caption{Results on Test Set (2022--2023)}
\label{tab:ablation}
\begin{tabular}{lccc}
\toprule
Model & Params & IoU $\uparrow$ & BF1 $\uparrow$ \\
\midrule
B1: Persistence        & N/A   & 0.160 & 0.128 \\
B2: Linear trend       & N/A   & 0.169 & 0.147 \\
\midrule
exp001: Image only     & 56.1M & 0.326 & 0.158 \\
exp002: + Climate      & 56.1M & \textbf{0.337} & 0.145 \\
exp003: + Climate + DEM& 56.1M & 0.331 & 0.123 \\
exp005: Climate only   & 0.66M & 0.320 & 0.135 \\
\bottomrule
\end{tabular}
\end{table}

\subsection{Comparison Against Traditional Baselines}
All deep learning models substantially outperform both
traditional baselines. The persistence baseline achieves
an IoU of 0.160, and the linear trend baseline achieves
0.169. The weakest deep learning model (exp005 climate-only
MLP, IoU 0.320) improves over linear trend by 89\% relative,
and the best model (exp002, IoU 0.337) improves by 99\%
relative. This is consistent with the models learning
meaningful temporal dynamics rather than simply
extrapolating static patterns.

\subsection{Effect of Climate Signals}
Adding ERA5 climate features (exp002) improves IoU from
0.326 to 0.337 (+3.4\% relative) over the image-only
baseline, suggesting that atmospheric forcing variables
carry predictive information beyond what Landsat imagery
alone provides. This improvement is consistent with the
hypothesis that climate signals precede visible glacier
boundary change.

\subsection{Effect of Terrain Features}
Adding DEM features to the climate-augmented model
(exp003, IoU 0.331) yields a slight regression compared
to exp002. One possible explanation is that static terrain
features (slope, aspect) are already implicitly encoded
in the spatial patterns of the Landsat imagery, making
their explicit inclusion redundant at the patch level.
This finding suggests that terrain features may require
more careful integration---for example, as conditioning
variables for the climate encoder rather than additive
spatial features.

\subsection{Climate-Only Baseline}
The climate-only MLP (exp005) achieves an IoU of
0.320---98\% of the image-only baseline (0.326) while
using 85$\times$ fewer parameters (661K vs.\ 56.1M).
ERA5 seasonal climate variables alone, with no access
to Landsat imagery, are nearly sufficient to predict
glacier boundary positions at the patch level. This
result is consistent with the hypothesis that climate
signals encode substantial predictive information about
glacier retreat that precedes visible imagery change.

\subsection{Per-Glacier Analysis}
\begin{figure}[t]
\centering
\includegraphics[width=\columnwidth]{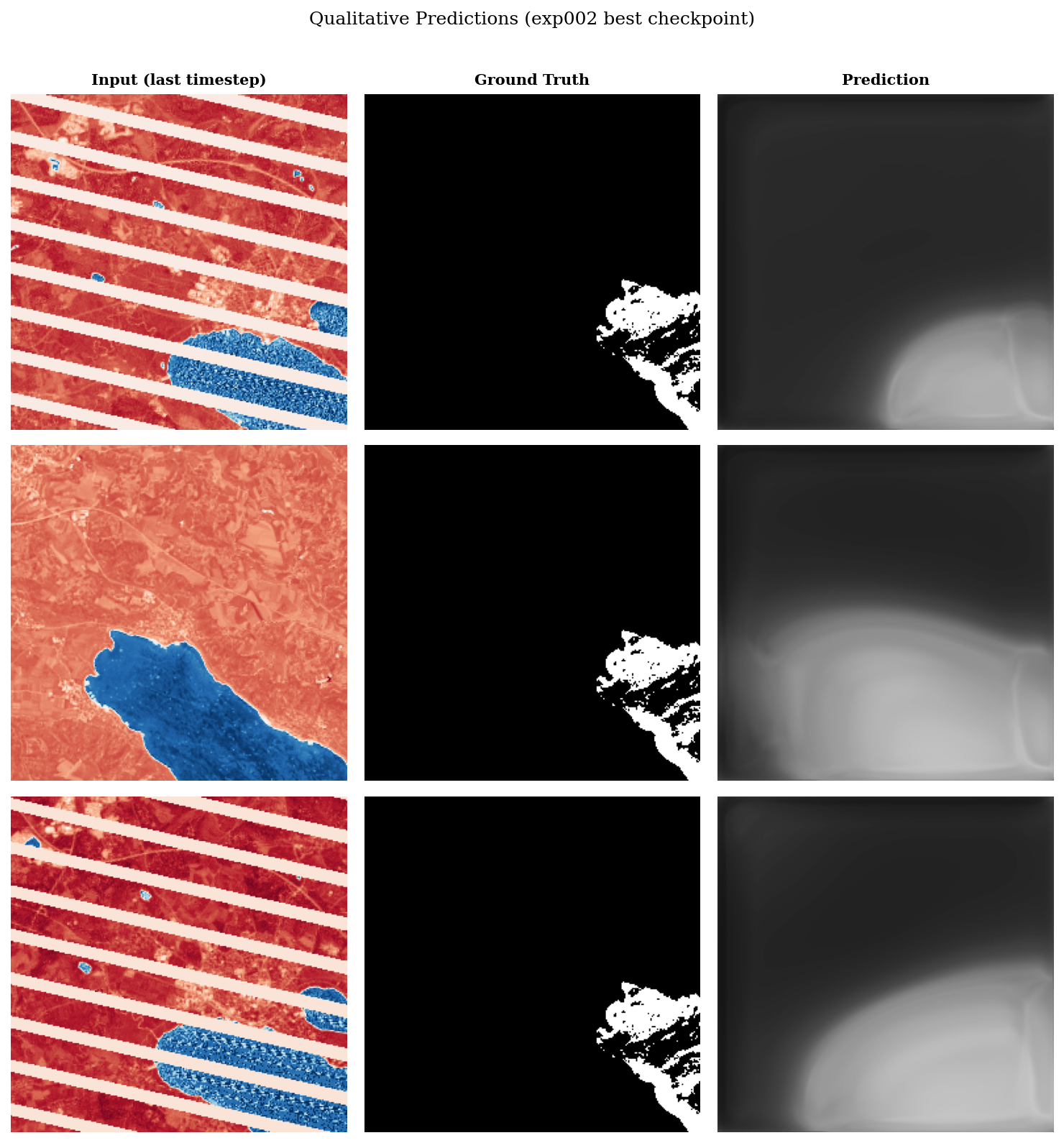}
\caption{Qualitative predictions from exp002. Each row shows
the last input timestep (NDSI channel), ground truth mask,
and predicted mask for three test patches.}
\label{fig:qualitative}
\end{figure}

Table~\ref{tab:perglacier} reports per-glacier IoU
for exp002 on the test set.
Fig.~\ref{fig:qualitative} shows qualitative predictions
for representative test patches. Columbia Glacier (Alaska)
achieves the highest IoU (0.500), likely reflecting
its large area and strong, consistent retreat signal.
Athabasca Glacier (IoU 0.045) performs poorly, consistent
with its small area (17.8~km$^2$) and the challenge of
predicting fine-scale boundaries from coarse ERA5
climate signals (~31km resolution). Gangotri Glacier
has no test sequences due to its scenes falling entirely
within the training years under the temporal split.

\begin{table}[h]
\centering
\caption{Per-Glacier IoU on Test Set (exp002)}
\label{tab:perglacier}
\begin{tabular}{lccc}
\toprule
Glacier & IoU & Std & $n$ patches \\
\midrule
Columbia  & 0.500 & 0.372 & 200 \\
Grey      & 0.250 & 0.201 & 200 \\
Aletsch   & 0.122 & 0.187 & 200 \\
Athabasca & 0.045 & 0.110 & 200 \\
Gangotri  & ---   & ---   & 0 (training split) \\
\bottomrule
\end{tabular}
\end{table}

\subsection{SHAP Climate Attribution}
\begin{figure}[t]
\centering
\includegraphics[width=\columnwidth]{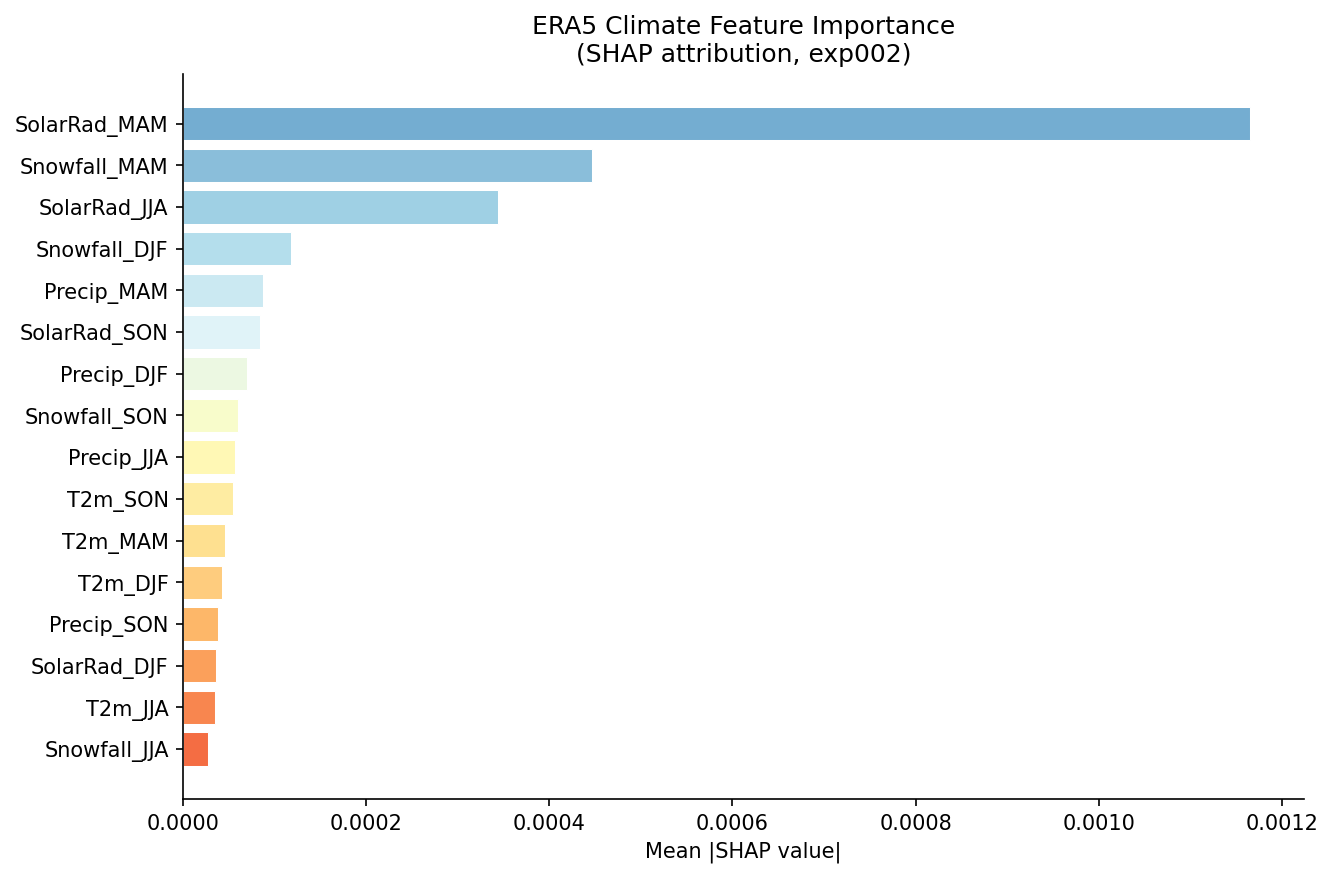}
\caption{SHAP feature importance ranking for all 16 ERA5
climate features. Spring solar radiation (SolarRad\_MAM)
is the highest-ranked driver of predicted glacier retreat.}
\label{fig:shap_summary}
\end{figure}

\begin{figure}[t]
\centering
\includegraphics[width=\columnwidth]{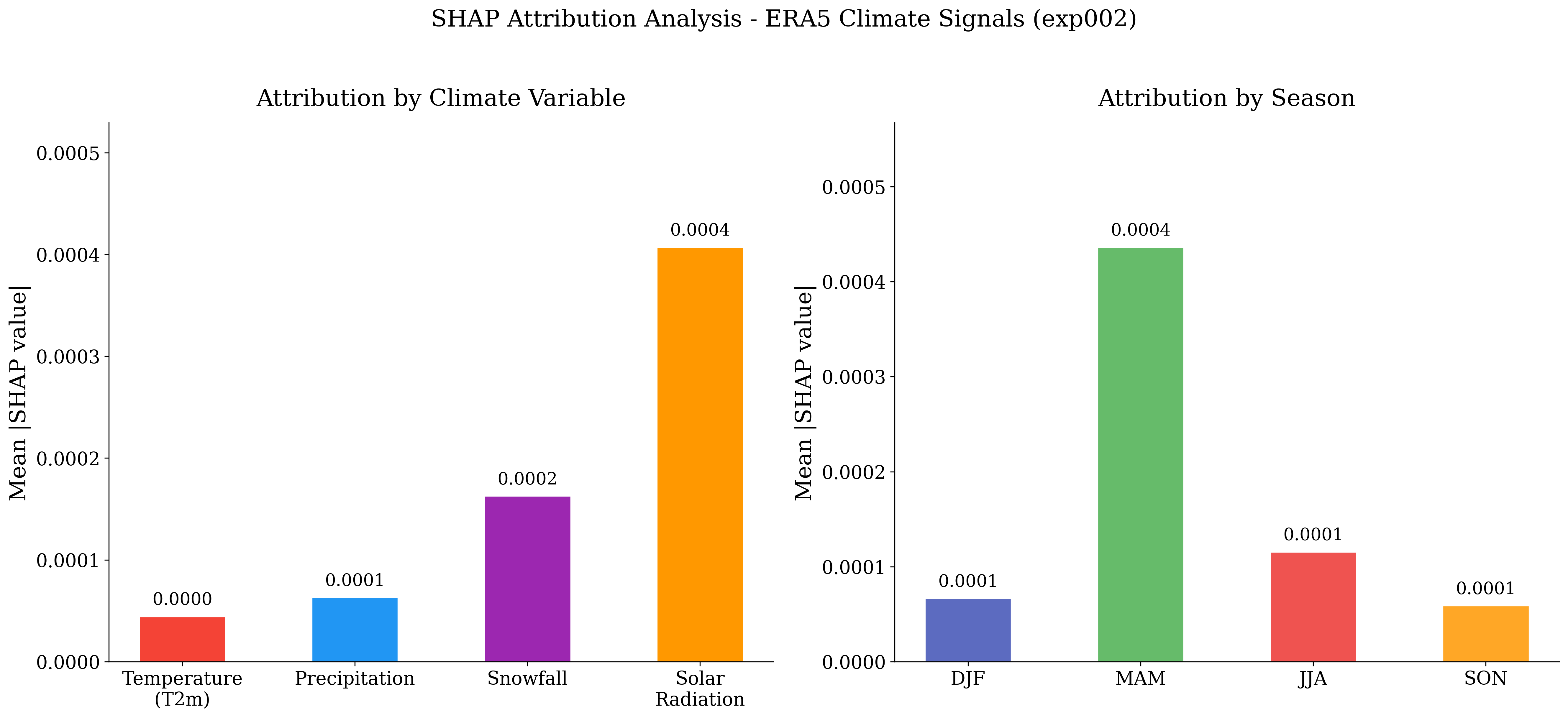}
\caption{SHAP attribution aggregated by climate variable
group (left) and season (right). Solar radiation contributes
most overall; spring (MAM) is the most predictive season.}
\label{fig:shap_seasonal}
\end{figure}

Table~\ref{tab:shap} reports mean absolute SHAP values
for the 16 ERA5 climate features, computed on 50 test
patches using the exp002 model. Figs.~\ref{fig:shap_summary}
and~\ref{fig:shap_seasonal} visualize the attribution
by feature and by variable group and season.

Spring solar radiation (SolarRad\_MAM) is the highest-ranked
driver, followed by spring snowfall (Snowfall\_MAM) and
summer solar radiation (SolarRad\_JJA). Aggregated by
variable group, solar radiation accounts for the largest
share of attribution, followed by snowfall and precipitation.
Temperature (T2m) contributes the least, which may reflect
its correlation with solar radiation variables from which
the model captures a similar signal.

\begin{table}[h]
\centering
\caption{Top ERA5 Climate Features by SHAP Attribution (exp002)}
\label{tab:shap}
\begin{tabular}{lcc}
\toprule
Feature & Mean $|$SHAP$|$ & Rank \\
\midrule
SolarRad\_MAM  & 0.0012 & 1 \\
Snowfall\_MAM  & 0.0004 & 2 \\
SolarRad\_JJA  & 0.0003 & 3 \\
Snowfall\_DJF  & 0.0001 & 4 \\
Precip\_MAM    & 0.0001 & 5 \\
\midrule
\multicolumn{3}{l}{\textit{Attribution by variable group (aggregated)}} \\
\midrule
Solar Radiation & 0.0004 & --- \\
Snowfall        & 0.0002 & --- \\
Precipitation   & 0.0001 & --- \\
Temperature     & 0.0000 & --- \\
\bottomrule
\end{tabular}
\end{table}

The prominence of spring solar radiation is physically
interpretable: the onset and intensity of the melt
season is primarily driven by incoming shortwave
radiation in spring (MAM), which influences how early
and how extensively the seasonal snowpack begins to
melt. This signal precedes the visible retreat of the
glacier terminus by weeks to months, consistent with
the temporal lag hypothesis motivating this work.
The secondary importance of spring snowfall reflects
its role in replenishing the snowpack and modulating
the effective melt season length.

\subsection{Discussion}
The absolute IoU values (0.320--0.337) are not directly
comparable to state-of-the-art glacier \textit{mapping}
models such as GlaViTU (IoU $> 0.85$)~\cite{glavitu2024},
because forecasting future boundaries from historical
sequences is inherently more uncertain than delineating
current boundaries from high-quality imagery. The
relevant comparisons are between experimental conditions
and traditional baselines, both of which GlacierCastAI
substantially exceeds.

The high variance in per-glacier performance (Columbia
0.500 vs.\ Athabasca 0.045) reveals a likely size
dependence: larger glaciers with stronger retreat signals
appear better predicted. This is consistent with the
spatial resolution mismatch between ERA5 climate data
(~31km) and the 30m Landsat patches --- fine-scale
climate variability relevant to small glaciers may be
smoothed out at ERA5 resolution. Future work should
investigate glacier-specific fine-tuning or area-weighted
loss functions to address this disparity.

A further limitation is that results reflect single
training runs per experimental condition, and multi-seed
evaluation has not been performed due to computational
constraints. Variance across seeds and bootstrap confidence
intervals represent important directions for strengthening
the statistical claims of this work.

\section{Conclusion}
\label{sec:conclusion}

This paper presented GlacierCastAI, which reframes glacier
boundary prediction as a multi-modal spatiotemporal
forecasting problem rather than a mapping problem.
Through a pre-registered ablation across four experimental
conditions, compared against traditional baselines and
interpreted via SHAP attribution, the results suggest that:

\begin{enumerate}
    \item All deep learning models substantially outperform
          persistence and linear trend baselines, improving
          IoU by 89--99\% relative, consistent with the
          models learning meaningful temporal dynamics.
    \item ERA5 climate signals improve glacier retreat
          forecasting IoU by 3.4\% over imagery alone,
          suggesting that atmospheric forcing provides
          predictive information not captured by Landsat
          sequences.
    \item A climate-only MLP with 661K parameters achieves
          98\% of image-only performance, suggesting that
          ERA5 seasonal variables are nearly sufficient
          for patch-level glacier boundary forecasting.
    \item DEM terrain features slightly reduce performance
          when added to imagery and climate inputs,
          suggesting redundancy with spatially encoded
          imagery features at the current patch scale.
    \item SHAP attribution identifies spring solar
          radiation (MAM) as the highest-ranked climate
          driver, followed by spring snowfall and summer
          solar radiation, consistent with the role of
          spring insolation in setting melt season
          trajectories.
\end{enumerate}

These results are consistent with the hypothesis that
climate signals precede visually detectable glacier
boundary change, supporting the feasibility of
climate-driven early warning systems for glacier retreat.
Future work will address multi-seed evaluation to
quantify result variance, per-glacier fine-tuning to
reduce the size-dependence of forecast accuracy,
integration of higher-resolution regional climate data
to improve small-glacier performance, and extension to
longer forecast horizons.

\bibliographystyle{IEEEtran}
\bibliography{references}

\end{document}